\documentclass[10pt, a4paper]{article}

\usepackage{lrec-coling2024} 
\usepackage{times}
\usepackage{latexsym}
\usepackage{graphicx}
\usepackage{booktabs}
\usepackage{array}

\usepackage[T1]{fontenc}
\usepackage{algorithmic}
\usepackage{algorithm}
\usepackage{inconsolata}
\usepackage{times}
\usepackage{amsmath}
\usepackage{amssymb}
\usepackage{subfigure}
\usepackage{multirow}

\title{TRELM: Towards Robust and Efficient Pre-training for Knowledge-Enhanced Language Models}

\name{Junbing Yan$^{1,2}$, Chengyu Wang$^2$, Taolin Zhang$^2$, Xiaofeng He$^1$, Jun Huang$^2$,\\ 
\bf \large Longtao Huang$^2$, Hui Xue$^2$, Wei Zhang$^1$
\thanks{Work done when Junbing Yan was doing an internship at Alibaba Group. Correspondence to Chengyu Wang and Wei Zhang.}}

\address{$^1$ School of Computer Science and Technology, East China Normal University \\ 
$^2$ Alibaba Group \\
 \texttt{\{junbingyan531,zhangwei.thu2011\}@gmail.com, hexf@cs.ecnu.edu.cn} \\
\texttt{\{chengyu.wcy,zhangtaolin.ztl,huangjun.hj,kaiyang.hlt,hui.xueh\}@alibaba-inc.com}\\}

\abstract{
KEPLMs are pre-trained models that utilize external knowledge to enhance language understanding. Previous language models facilitated knowledge acquisition by incorporating knowledge-related pre-training tasks learned from relation triples in knowledge graphs. However, these models do not prioritize learning embeddings for entity-related tokens. Moreover, updating the entire set of parameters in KEPLMs is computationally demanding. This paper introduces \textbf{TRELM}, a Robust and Efficient Pre-training framework for Knowledge-Enhanced Language Models. We observe that entities in text corpora usually  follow the long-tail distribution, where the representations of some entities are suboptimally optimized and hinder the pre-training process for KEPLMs. To tackle this, we employ a robust approach to inject knowledge triples and employ a knowledge-augmented memory bank to capture valuable information. Furthermore, updating a small subset of neurons in the feed-forward networks (FFNs) that store factual knowledge is both sufficient and efficient. Specifically, we utilize dynamic knowledge routing to identify knowledge paths in FFNs and selectively update parameters during pre-training. Experimental results show that TRELM reduces pre-training time by at least 50\% and outperforms other KEPLMs in knowledge probing tasks and multiple knowledge-aware language understanding tasks.
 \\ \newline  }

\begin{document}

\maketitleabstract

\section{Introduction}
\label{intro}
\begin{figure*}
\centering
\includegraphics[width=\textwidth]{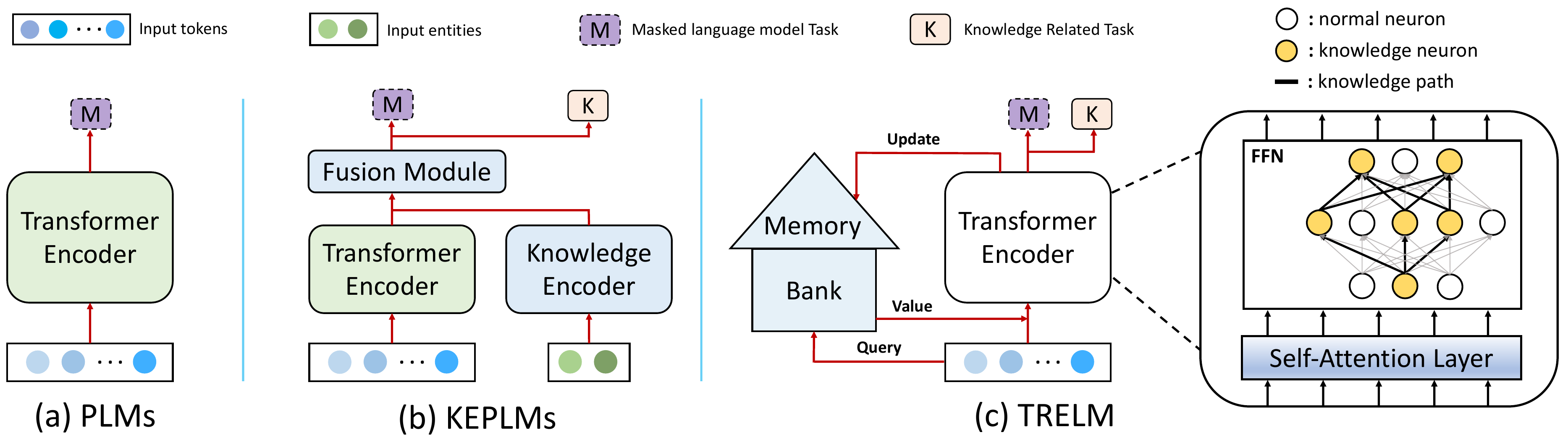}
\caption{Comparison between TRELM and other models. (a) Plain PLMs usually utilize masked language modeling as the pre-training objective. (b) Some KEPLMs utilize external knowledge sources (e.g., KGs) and design knowledge-aware tasks which need additional knowledge encoders. (c) During pre-training, TRELM uses a BERT-style shared encoder and a knowledge-augmented memory bank to inject factual knowledge. Moreover, we only need to update partial FFN parameters in Transformer blocks with a dynamic knowledge routing method.}
\label{motivation}
\end{figure*}

Pre-trained language models (PLMs) such as BERT \citep{bert} and RoBERTa \citep{roberta} learn language representations from large-scale text corpora and significantly improve the performance of various NLP tasks \cite{xu2021syntax,chang2021convolutions}. Yet, they often lack methods for incorporating external knowledge for language understanding (\citealp{colon2021combining}; \citealp{cui2021commonsense}). Since knowledge graphs (KGs) can provide rich structured knowledge facts (\citealp{knowledge-bases-lstms}; \citealp{Adaptive-knowledge-sharing}; \citealp{Neural-knowledge-acquisition}), the performance of PLMs can be enhanced by injecting external knowledge triples from KGs, known as Knowledge-Enhanced PLMs (KEPLMs). KEPLMs (\citealp{zhang2019ernie}; \citealp{kepler}; \citealp{colake}; \citealp{zhang2022dkplm}) incorporate knowledge-related tasks, such as denoising entity auto-encoder (dEA) and knowledge embedding learning, to facilitate knowledge understanding in the models. Figure~\ref{motivation} summarizes the distinctions between PLMs without external knowledge integration and KEPLMs.

Despite the success of KEPLMs, two main problems still remain.
(1) Most of the previous KEPLMs indiscriminately inject knowledge into PLMs, which can introduce noisy knowledge such as redundant or irrelevant information, potentially degrading model performance~\citep{knowbert}.
These methods~\cite{zhang2019ernie, k-bert, kepler, colake} inject corresponding knowledge triples or pre-trained knowledge embeddings into each entity in the context. However, some entities appear frequently in texts, leading to redundant knowledge injection.
Irrelevant knowledge arises when some entities or their corresponding sub-graphs have little connection to the meanings of the underlying sentences; hence, they contribute minimally to the improvement of model performance.
(2) Some methods modify model backbones with additional knowledge encoders, leading to inflexibility~\cite{zhang2022dkplm}. Furthermore, optimizing these encoders can adversely affect the model's computational efficiency.
Recently, some works (\citealp{aximomatic-attribution}; \citealp{attattr}) use attribution to explain the mechanism of the Transformer. Most regard the self-attention layers as key-value pairs, while the study~\citep{dai2022knowledge} views feed-forward networks (FFNs) as key-value memories and points out that some neurons in FFNs relate to knowledge expressions, motivating us to explore a similar spirit in KEPLMs.

In this paper, we present our contributions in the form of our novel KEPLM training paradigm, namely TRELM, which enables the pre-training of more robust and efficient KEPLMs. To address the issue of excessive knowledge noise introduction, we propose identifying important entities as targets for knowledge injection. To facilitate the learning of improved representations, we construct a knowledge-augmented memory bank, which is vital in guiding the pre-training process and expediting convergence. Moreover, to optimize computational resource utilization, we introduce a technique called dynamic knowledge routing. This involves selective parameter updates within Transformer blocks. By identifying knowledge paths based on knowledge attribution, we enable partial updates of model parameters, focusing on the FFNs. Consequently, this results in a more efficient utilization of computing resources.

We conduct extensive experiments to verify the robustness and effectiveness of our TRELM framework over multiple NLP tasks. Our results show that TRELM outperforms strong baselines in knowledge-related tasks, including knowledge probing (LAMA) \citep{lama}, relation extraction, and entity typing. The pre-training time is also significantly reduced by over 50\%.
In summary, the contributions of this paper are as follows:~\footnote{Source codes will be publicly available in the EasyNLP framework~\cite{DBLP:conf/emnlp/WangQZLLWWHL22}. URL:~\url{https://github.com/alibaba/EasyNLP}}

\begin{itemize}
\item \textbf{New Pre-training Paradigm.} We introduce a more robust and efficient knowledge-enhanced pre-training paradigm (TRELM).

\item \textbf{Knowledge-augmented Memory Bank.} We detect important entities in pre-training corpora and construct a knowledge-augmented memory bank, which guides the pre-training process and accelerates convergence.

\item \textbf{Dynamic Knowledge Routing.} We propose a novel knowledge routing method that dynamically finds knowledge paths in FFNs and selectively updates model parameters.



\item \textbf{Comprehensive Experiments.} We conduct extensive experiments and case studies to show the effectiveness and robustness of TRELM over various NLP tasks.

\end{itemize}

\section{Related Work}
In this section, we survey literature relevant to our study, encompassing three primary domains: KEPLMs, attribution methods in Transformer architectures, and the application of attribution to KEPLMs.

\begin{figure*}[t]
\centering
\includegraphics[height=10.5cm]{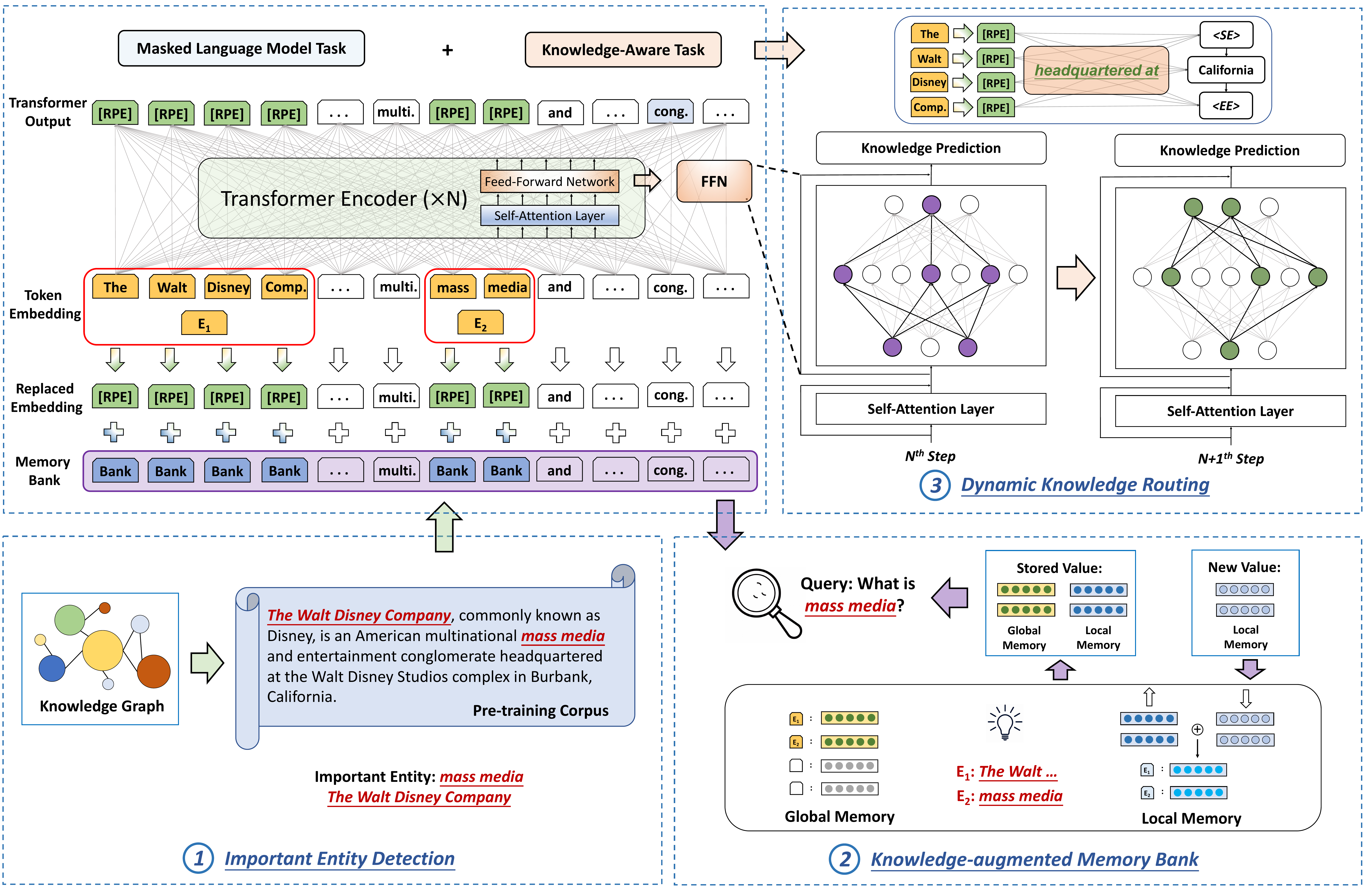}
\caption{Model overview. 
(1) \textbf{Input:} Detecting important entities and long-tail words to reduce the knowledge noises. (2) \textbf{Knowledge-augmented Memory Bank:} Querying the important knowledge learned previously through a ``cheat sheet'' that contains semantic information of entities and words. (3) \textbf{Dynamic Knowledge Routing:} Finding the knowledge paths related to the knowledge-aware task, and selectively update the model's parameters.} 
\label{model_overview}
\end{figure*}

\subsubsection{KEPLMs}
KEPLMs incorporate external knowledge to enhance language understanding abilities of PLMs (\citealp{baidu-ernie}; \citealp{zhang2019ernie}; \citealp{knowbert}; \citealp{xiong2019pretrained}; \citealp{k-adapter}; \citealp{k-bert}; \citealp{kepler}; \citealp{colake}; \citealp{zhang2022dkplm}; \citealp{ye2022ontology}; \citealp{yu2022jaket}; \citealp{DBLP:conf/emnlp/ZhangDWWWLHLH22}; \citealp{DBLP:conf/cikm/ZhangC0LLQTHH21}). 
For instance, ERNIE-Baidu \citep{baidu-ernie} introduces entity and phrase level masking strategies to capture semantic information, while ERNIE-THU \citep{zhang2019ernie} integrates entity embeddings into contextual representations using knowledge encoders. K-BERT \citep{k-bert} and CoLAKE \citep{colake} exploit knowledge graphs (KGs) to augment the language model with graph structures, and DKPLM \citep{zhang2022dkplm} employs shared encoders to unify texts and entities within a single semantic space.
Despite these advancements, KEPLMs face ongoing challenges that limit their effectiveness and versatility. 
These challenges form the basis of our study and drive our exploration into novel methods for enhancing PLMs with external knowledge.

\subsubsection{Attribution Methods in Transformers}
Integrated gradients, a technique for attributing the model's output to its input features, has been increasingly adopted (\citealp{attattr}; \citealp{dai2022knowledge}). For instance, \citet{attattr} applied integrated gradients to the self-attention mechanism, elucidating the importance of specific attention heads in the model's computations. 
More recent discussions by \citet{wu2019pay} and \citet{dong2021attention} have expanded the focus beyond self-attention, highlighting the significant role of FFNs within Transformers. \citet{dai2022knowledge} employed integrated gradients to investigate the ``knowledge neurons'' in FFNs, providing insights into how these models process and store factual knowledge.

\subsubsection{Attribution for KEPLMs}
In the realm of KEPLMs, the challenge of filtering out knowledge noise has emerged as a critical concern. Several studies (\citealp{knowbert}; \citealp{lama}; \citealp{cao2021kb}; \citealp{colake}; \citealp{klmo}; \citealp{zhang2022dkplm}; \citealp{kepler}; \citealp{DBLP:conf/emnlp/ZhangX0DCQCHQ23}) have demonstrated that the presence of knowledge noise can significantly impair model performance. Our research posits that the concept of knowledge paths, which are sequences of knowledge neurons within FFN layers of a Transformer, is instrumental to the effectiveness of KEPLMs. 

\section{TRELM: The Proposed Framework}

We first state some basic notations.
Denote an input token sequence as $x=(x_1,\cdots, x_i,\cdots, x_n)$, where $n$ is the sequence length.
The hidden representations of input tokens are denoted as $(h_1, h_2, \ldots, h_n)$ and $h_i \in \mathbb{R}^{d_1}$, where $d_1$ is the dimension of the representations.
Furthermore, a knowledge graph is denoted as $\mathcal{G}=(\mathcal{E}, \mathcal{R})$. Here, $\mathcal{E}$ and $\mathcal{R}$ are the sets of entities and relation triples, respectively.
In the KG, a relational knowledge triple $(e_h, r, e_t)$ comprises the head entity $e_h$, the relation $r$, and the tail entity $e_t$.
The overall framework of TRELM is illustrated in Figure \ref{model_overview}. 
We aim to address three key research questions:
\begin{itemize}
    \item \textbf{RQ1}: How can we select useful positions and design more effective techniques for knowledge injection?
    \item \textbf{RQ2}: How can we ensure that our model retains the injected knowledge?
    \item \textbf{RQ3}: How can parameters of TRELM be efficiently updated during the pre-training knowledge learning process, while preserving downstream task performance?
\end{itemize}


\subsection{Noise-aware Knowledge Injection}
\label{know_injection}
\noindent\textbf{Important Entity Infusion.}
As shown by \citealp{lama}, \citealp{broscheit2019investigating}, \citealp{wang2020language}, and \citealp{cao2021kb}, the semantics of high-frequency and common relation triples are already captured by plain PLMs.
In this section, we aim to detect important entities for robust knowledge injection and knowledge noise reduction. In our work, selecting important entity positions in pre-training sentences for knowledge injection is vital. Inspired by \citet{zhang2022dkplm}, we define the \emph{Semantic Importance} (SI) score of the entity $e$ in the sentence as $SI(e)$, indicating the semantic similarity between the representation of the original sentence and that of the sentence with $e$ being replaced. We select our desired entities with high $SI(e)$ scores as target injection positions:
\begin{equation}
    SI(e) = \frac{\left \| h_o \right \| \cdot \left \| h_{rep} \right \|}{h_o \cdot h_{rep}}.
\end{equation}

\noindent\textbf{Contrastive Knowledge Assessing.}
Knowing where to inject knowledge is insufficient, as knowledge constraints are only applied to the input layer. It is also necessary to verify whether the model truly acquires the knowledge. For model optimization, in addition to the Masked Language Modeling (MLM) task \citep{bert}, we propose \emph{Contrastive Knowledge Assessing} (CKA) as an additional pre-training task. 

The basic idea is that, given the representations of a head entity in the pre-training sentence and a relation at the input layer, the model needs to determine at the output layer whether it detects whether a given entity is the correct tail entity or not, and vice versa. Specifically, for the predicted $i$-th tokens of the tail entity $h_d^{i}$ \cite{zhang2022dkplm}, we employ deep contrastive learning to encourage the model to capture the knowledge. Let $f(h_d^{i},\cdot)$ be a matching function between $h_d^{i}$ and a result token. The \emph{token-level} CKA loss function is as follows:
\begin{equation}
\mathcal{L} = -\log \frac{\exp(f(h_d^{i},y_i))}{\exp(f(h_d^{i},y_i)) + \sum_{y_i^{'}\sim Q(y_i)}\exp (f(h_d^{i},y_i^{'}))}
\end{equation}
where $y_i$ is the ground-truth token, and $y_i^{'}$ is a negative token sampled from a negative sampling function $Q(y_i)$. Hence, the total loss function of TRELM is:
\begin{equation}
\label{eq:loss}
\mathcal{L}_{\mathrm{total}}=\theta\cdot\mathcal{L}_{\mathrm{MLM}}+(1-\theta)\cdot\mathcal{L}_{\mathrm{CKA}}
\end{equation}
where $\theta$ is the hyper-parameter, and $\mathcal{L}_{\mathrm{CKA}}$ is the contrastive loss with respect to target entities.

\subsection{Enhancing Representations with Knowledge-augmented Memory Bank}

We have explored knowledge injection for important entities. Yet, since entities in the corpus typically follow a ``long-tail'' distribution~\cite{TNF, zhang2022dkplm}, some representations can still be poorly optimized.
Here, we further construct a \emph{Knowledge-augmented Memory Bank} (KMB), which acts as a ``cheat sheet'' to ensure the model consistently captures important knowledge learned previously.

\noindent\textbf{KMB Construction with Global and Local Memory Enhancement.}
\citet{TNF} discovered that learning representations for rare tokens during pre-training is challenging. It is reasonable to extend this hypothesis to knowledge-enhanced learning. However, their study focuses only on the local memory of infrequent tokens without considering the global memory of tokens.
When encountering an important entity in a sentence, we can treat the contextual representations of its surrounding words as its ``local memory.''
In detail, we construct a KMB $\mathcal{M}$.
For an entity $e$ present in both sentence $x$ and $\mathcal{M}$, we denote the span boundary of $e$ in $x$ as $(l,r)$, with $l$ and $r$ being the starting and ending positions, respectively. The ``local memory'' of $e$ for $x$ is defined as:
\begin{equation}\label{eq:takinglocalmomery}
\mathcal{M}_{local}^{(e,x)}=\frac{1}{2k+r-l}\sum_{i=l-k}^{r+k}\mathbf{h}_i,
\end{equation}
where $\mathbf{h}_i \in \mathbb{R}^{d_1}$ is the output at position $i$ of the Transformer encoder, serving as the contextual representation of $x$. Here, $k$ is half the window size and controls the number of surrounding tokens.

Since entity $e$ may appear multiple times in the pre-training corpus, in $\mathcal{M}$, the ``local memory'' for entity $e$ in KMB (denoted as $\mathcal{M}_{local}^{(e)}$) is updated using a moving average of every $\mathcal{M}_{local}^{(e,x)}$ that we obtain. We initialize $\mathcal{M}_{local}^{(e)}$ using the pre-trained embeddings of RoBERTa \citep{roberta}. Therefore, at any occurrence of entity $e$ during pre-training, its contextual information from all previous occurrences can be leveraged. We update $\mathcal{M}_{local}^{(e)}$ as:
\begin{equation}\label{eq:update}
\mathcal{M}_{local}^{(e)} \leftarrow (1-\gamma)\cdot\mathcal{M}_{local}^{(e)}+\gamma \cdot\mathcal{M}_{local}^{(e,x)}
\end{equation}
where $\gamma\in(0,1)$ is the discount factor. Since the local memory contains localized information subject to isolation, we propose aggregating representations of $e$ across multiple contexts as the ``global memory''. Let $\mathcal{T}^{(m)}$ be the collection of contexts involving entity $e$, i.e., $\mathcal{T}^{(m)} = \{\mathcal{T}_{n} \vert n \in \{1,\cdots,N\}, e \in \mathcal{T}_{n} \}$, and let $\mathbf{h}_{cls}$ be the output for the special \texttt{<cls>} classification token by the last Transformer layer. The ``global memory'' of entity $e$ can be denoted as follows:
\begin{equation}\label{eq:takingglobalmomery}
\mathcal{M}_{global}^{(e)}=\frac{1}{\lvert \mathcal{T}^{(m)} \rvert}\sum_{\mathcal{T}_{n} \in \mathcal{T}^{(m)}}\mathbf{h}_{cls}.
\end{equation}

\noindent\textbf{Leveraging KMB for Pre-training.} 
We leverage the stored representations of entities in KMB as part of the input to the encoder. For any token sequence $x=\{x_1,\cdots, x_i,\cdots, x_n\}$, we first identify all important entities $e$ appearing in $x$. Assuming that there are $n$ important entities, they are denoted as $\{(e_i,l_i,r_i)\}_{i=1}^n$ where $(l_i, r_i)$ are the boundaries of $e_i$ in $x$ at the $i$-th position respectively. If $l_i \leq p \leq r_i$, at position $p$, the input embeddings to the model are defined as follows:
\begin{equation}
\label{eq:input}
\mathcal{I}_p =(1-\lambda)\cdot{h_{e_i}}+ \frac{\lambda}{2} \cdot(\mathcal{M}_{local}^{(e_i)}+\mathcal{M}_{global}^{(e_i)})
\end{equation}
Otherwise, we have: $\mathcal{I}_p = \mathcal{E}_p$
where $\mathcal{E}_p$ is the token embedding at position $p$, $h_{e_i}$ is the knowledge injection embedding for $e_i$, and $\lambda$ is a hyper-parameter controlling the degree to which our KEPLM relies on KMB for contextual representations of important entities. We empirically set $\lambda$ to 0.5 initially. To mitigate bias between pre-training and fine-tuning (which does not involve KMB), $\lambda$ gradually decays to 0 towards the end of pre-training, i.e.:
\begin{equation}
   \lambda_q=\frac{1}{\beta^q} \cdot\lambda, \quad   q = 0, 1, 2, \cdots
\end{equation}
where $\beta$ is a hyper-parameter that controls the decay rate of $\lambda$, and $q$ is the pre-training epoch.

\begin{table*}[t]
\centering
\begin{small}
\begin{tabular}{c|cc|ccccc|cc}
\toprule
\multirow{2}*{Datasets} & \multicolumn{2}{c}{PLMs} & \multicolumn{7}{c}{KEPLMs} \\ \cmidrule(r){2-3} \cmidrule(r){4-10}
~ & ELMo & RoBERTa &  CoLAKE & KEPLER & DKPLM & KP-PLM & KALM & TRELM & $\bigtriangleup$ 	 \\
\midrule
Google-RE & 2.2\%  & 5.3\%  & 9.5\%  & 7.3\% & 10.8\% & 11.0\% & 10.2\% & \textbf{11.5}\% & +0.5\% \\
UHN-Google-RE & 2.3\%  & 2.2\% & 4.9\%  & 4.1\% &  5.4\% & 5.6\%& 5.2\% & \textbf{5.9}\% & +0.3\%  \\ \midrule
T-REx & 0.2\%   & 24.7\% & 28.8\%  & 24.6\% & 32.0\% & 32.3\% & 29.8\%& \textbf{33.0}\% & +0.7\%  \\
UHN-T-REx & 0.2\%  & 17.0\% & 20.4\% & 17.1\%  & 22.9\% & 22.5\% & 22.6\% & \textbf{23.3}\% & +0.4\% \\
\bottomrule
\end{tabular}
\end{small}
\caption{The performance on knowledge probing. $\bigtriangleup$ represents the absolute improvements over the best results of existing KEPLMs compared to our model.}
\label{knowledge_probing_result_RoBERTa}
\end{table*}

\begin{table}[t]
\centering
\begin{small}
\begin{tabular}{c|ccc}
\toprule
Datasets & BERT & TRELM$_{BERT}$ & $\bigtriangleup$ 	 \\
\midrule
Google-RE & 11.4\%  & \textbf{15.3}\%  & +3.9\% \\
UHN-Google-RE & 5.7\%  & \textbf{9.8}\%  & +4.1\%  \\ \midrule
T-REx & 32.5\% & \textbf{36.7}\%  & +4.2\%  \\
UHN-T-REx & 23.3\%  & \textbf{27.9}\% & +4.6\% \\
\bottomrule
\end{tabular}
\end{small}
\caption{The performance on knowledge probing based on BERT. $\bigtriangleup$ represents the absolute improvements over BERT compared to TRELM.}
\label{knowledge_probing_result_BERT}
\end{table}

\subsection{Learning with Dynamic Knowledge Paths} 
After determining the model inputs and outputs, we proceed with the parameter optimization process. Building upon the hypothesis by~\citet{dai2022knowledge}, which suggests that factual knowledge is stored in the FFN layers of Transformers, we introduce a \emph{dynamic knowledge routing} algorithm to identify critical knowledge paths for TRELM updates during knowledge acquisition. Given an input sequence $x$, we define $\operatorname{P}_x(\hat{v}^{(l)}_{i})$ as the probability of producing the correct response according to the knowledge assessing objective:
\begin{equation}
    \operatorname{P}_x(\hat{v}^{(l)}_{i}) = p(y^* | x, v^{(l)}_{i}=\hat{v}^{(l)}_{i})
\end{equation}
where $p$ represents the Sampled SoftMax function; $y^*$ is the correct response; 
$v^{(l)}_{i}$ is the $i$-th neuron in the $l$-th FFN layer; and
$\hat{v}^{(l)}_{i}$ is a specific value of $v^{(l)}_{i}$.
As $v^{(l)}_{i}$ varies from 0 to its upper bound $\overline{v}^{(l)}_{i}$, we calculate the neuron's attribution score by integrating the gradients of ${P}_x(\alpha {v}^{(l)}_{i})$:
\begin{equation}
\operatorname{Attr}(v^{(l)}_{i}) = \overline{v}^{(l)}_{i} \int_{\alpha=0}^{1} \frac{\partial \operatorname{P}_x(\alpha \overline{v}^{(l)}_{i})}{\partial v^{(l)}_{i}} d \alpha.
\end{equation}
The attribution score, ${Attr}(v^{(l)}_{i})$, quantifies the impact of $v^{(l)}_{i}$ on the output probabilities using integrated gradients as $\alpha$ spans from 0 to 1. However, directly calculating the continuous integral is challenging; thus, we use the Riemann approximation:
\begin{equation}
\Tilde{\operatorname{Attr}}(v^{(l)}_{i}) = \frac{\overline{v}^{(l)}_{i}}{m} \sum_{k=1}^{m} \frac{\partial \operatorname{P}_x(\frac{k}{m} \overline{v}^{(l)}_{i})}{\partial v^{(l)}_{i}}
\end{equation}
where $m$ is set to 20 based on empirical testing.

Neurons with high $\operatorname{Attr}(v^{(l)}_{i})$ scores are indicative of a strong association with the understanding of knowledge within FFN layers. We define the knowledge path in the $\mathcal{T}$-th FFN layer as the sequence:
\begin{equation}
(\mathcal{T}_{{v}^{(i)}_{input}}\!\rightarrow \mathcal{T}_{{v}^{(j)}_{inter}}\!\rightarrow \mathcal{T}_{{v}^{(k)}_{output}})
\end{equation}
where $\mathcal{T}_{{v}^{(i)}_{input}}$, $\mathcal{T}_{{v}^{(j)}_{inter}}$, and $\mathcal{T}_{{v}^{(k)}_{output}}$ represent the $i$-th, $j$-th, and $k$-th neurons associated with knowledge in the FFN's input, intermediate, and output layers, respectively. These connections are crucial to the factual knowledge present in KEPLM. By selectively updating the model parameters based on the gradients of these knowledge paths, we can significantly reduce the computational cost of pre-training. Our experiments confirm that this technique not only accelerates pre-training convergence but also improves the model's understanding capabilities.

\noindent\textbf{Remarks.} During pre-training, we efficiently identify knowledge paths for each batch in parallel by utilizing distinct knowledge decoding labels. Although the detection of knowledge paths adds some overhead, the reduction in back-propagation time during model pre-training far outweighs this initial cost. This streamlined approach not only enhances efficiency but also contributes to
effectiveness in capturing relevant knowledge.


\subsection{Summarization of Pre-training Process}
We provide a summary of the entire pre-training procedure below.

\noindent\textbf{Input.} We identify important entities and long-tail words throughout the corpus. Entity embeddings are replaced with those generated by the KG embedding algorithm discussed in Section~\ref{know_injection}. Embeddings of important entities and long-tail words are then updated in the Knowledge-augmented Memory Bank (KMB).

\noindent\textbf{Forward Pass.} During each FFN layer, we calculate the attribution scores for the neurons. These scores allow us to evaluate the importance level of knowledge neurons and establish knowledge paths. Following a forward pass, KMB values are updated based on the output from the model's final Transformer layer.

\noindent\textbf{Back Propagation.} In the final step, we selectively update the parameters along the identified knowledge paths during back propagation, focusing the training on the most relevant aspects of the model's knowledge representation.

\begin{table*}[thbp]
  \centering\small
  \begin{tabular}{l|cccccccc|c}
  \toprule
  Model     & MNLI (m/mm) & QQP   & QNLI & SST-2 & CoLA & STS-B & MRPC & RTE  & AVG. \\ \midrule
  RoBERTa   & 87.5 / 87.3 & 91.9  & \textbf{92.8} & 94.8  & 63.6 & 91.2  & 90.2 & 78.7 & 86.4 \\
  KEPLER    & 87.2 / 86.5 & 91.5  & 92.4 & 94.4  & 62.3 & 89.4  & 89.3 & 70.8 & 84.9 \\
  CoLAKE     & 87.4 / 87.2 & 92.0  & 92.4 & 94.6  & 63.4 & 90.8  & 90.9 & 77.9 & 86.3 \\ \midrule
 TRELM      & \textbf{87.9} / \textbf{87.3} & \textbf{92.2}  & 92.6 & \textbf{94.9}  & \textbf{63.9} & \textbf{91.5}  & \textbf{91.2} & \textbf{79.1} & \textbf{86.7} \\ \bottomrule
  \end{tabular}
  \caption{GLUE results on dev set. KEPLER, CoLAKE and TRELM are initialized with RoBERTa$_\mathrm{BASE}$.}
  \label{tab:glue}
\end{table*}

\begin{table}
\centering
\begin{small}
\begin{tabular}{l | ccc}
\toprule
Model & Precision & Recall & F1 \\
\midrule
BERT & 76.4$\pm$1.2 & 71.0$\pm$1.4 & 73.6$\pm$1.3 \\
RoBERTa & 77.4$\pm$1.8 & 73.6$\pm$1.7 & 75.4$\pm$1.8 \\ 
\midrule
ERNIE$_{BERT}$ & 78.4$\pm$1.9 &72.9$\pm$1.7 &75.6$\pm$1.9 \\
ERNIE$_{RoBERTa}$ & 80.3$\pm$1.5 &70.2$\pm$1.7 &74.9$\pm$1.4 \\
KnowBERT & 77.9$\pm$1.3 &71.2$\pm$1.5&74.4$\pm$1.3 \\
KEPLER$_{WiKi}$ & 77.8$\pm$2.0& 74.6$\pm$1.9& 76.2$\pm$1.8  \\
CoLAKE & 77.0$\pm$1.6 &75.7$\pm$1.7&76.4$\pm$1.5 \\ 
DKPLM & 79.2$\pm$1.3 & 75.9$\pm$1.2& 77.5$\pm$1.2 \\ 
KP-PLM & \textbf{80.8}$\pm$1.7& 75.1$\pm$1.6 & 77.8$\pm$1.7\\ 
KALM & 78.9$\pm$1.5 & 75.3$\pm$1.6& 77.1$\pm$1.6\\
\midrule
TRELM  & 80.2$\pm$1.3 &\textbf{76.0}$\pm$1.4&\textbf{78.0}$\pm$1.2 \\
\bottomrule
\end{tabular}
\end{small}
\caption{Model performance on Open Entity (\%).}
\label{open_entity_result}
\end{table}

\section{Experiments}
In this section, we comprehensively evaluate the effectiveness of TRELM and compare it against state-of-the-art approaches. 

\subsection{Experimental Setup}
\noindent\textbf{Pre-training Data.}
For pre-training TRELM, we utilize the English Wikipedia dated 2020/03/01\footnote{https://dumps.wikimedia.org/enwiki/} as our data source. We align entities in the pre-training texts, recognized by entity linking tools such as TAGME \citep{tagme}, with the Wikidata5M \citep{kepler} knowledge graph. Wikidata5M provides a large-scale dataset that includes relation triples and entity description texts. Additional pre-processing and filtering steps are consistent with those used by ERNIE-THU \citep{zhang2019ernie}. As a result, our KG comprises 3,085,345 entities and 822 relation types, and we have prepared 26 million text sequences.

\noindent\textbf{Baselines.}
We compare TRELM with the following state-of-the-art KEPLM approaches:
\begin{enumerate}
    \item \textbf{ERNIE-THU}~\citep{zhang2019ernie}: Integrates knowledge embeddings by introducing a new pre-training objective that aligns mentions with knowledge entities.
    \item \textbf{KnowBERT}~\citep{knowbert}: Enhances language representations with structured knowledge through knowledge attention.
    \item \textbf{KEPLER}~\citep{kepler}: Encodes entities alongside text within Transformer blocks to create a joint semantic space.
    \item \textbf{CoLAKE}~\citep{colake}: Utilizes a knowledge graph and adjacency matrices to guide the information flow.
    \item \textbf{DKPLM}~\citep{zhang2022dkplm}: Detects long-tail entities and uses a shared encoder for the injection of knowledge triples.
    \item \textbf{KP-PLM}~\citep{KP-PLM}: Uses continuous prompts and introduces two knowledge-aware self-supervised tasks for pre-training.
    \item \textbf{KALM}~\citep{KALM}: Incorporates external knowledge into three levels of document contexts for language understanding.
\end{enumerate}

\subsection{Knowledge-aware Tasks}
TRELM was evaluated on three knowledge-aware tasks: knowledge probing (in the zero-shot setting), relation extraction, and entity typing. Due to space constraints, the primary experiments utilized RoBERTa$_{BASE}$ as the underlying architecture. The results demonstrate TRELM's transferability to larger models.

\noindent\textbf{Knowledge Probing}: The LAMA \citep{lama} probes use cloze-style tasks (e.g., "Arroyo died at \texttt{[MASK]} in 1551.") to assess whether PLMs encapsulate factual knowledge. The LAMA-UHN \citep{lamauhn} subset presents a more challenging set of questions by removing samples that are easier to answer. TRELM's performance on these tasks was quantified using macro-averaged mean precision (P@1), which gauges the model's ability to retrieve correct facts accurately.

The results for the LAMA and LAMA-UHN tasks can be found in Table~\ref{knowledge_probing_result_RoBERTa} and Table~\ref{knowledge_probing_result_BERT}. BERT-based models were separated from RoBERTa-based ones due to the significantly smaller vocabulary size of BERT, as per insights from \cite{k-adapter}. The primary findings are as follows: 
(1) Our model, built on RoBERTa-base, attains state-of-the-art results across four datasets. 
(2) To ensure a balanced comparison, TRELM was also trained on the BERT-base model. As displayed in Table~\ref{knowledge_probing_result_BERT}, TRELM significantly surpasses BERT-base, with an average improvement of +4.2\%, reinforcing that TRELM is an effective pre-training framework adaptable to various architectures.

\noindent\textbf{Entity Typing}: This task requires predicting the semantic type of a specified entity within a given context. To ensure a fair comparison, we adhere to the training settings used in \cite{zhang2022dkplm} and evaluate TRELM on the Open Entity dataset \citep{openentity}. Consistent with prior studies, we report micro-averaged precision, recall, and F1 metrics. As Table~\ref{open_entity_result} shows, KEPLMs generally outperform plain PLMs due to additional knowledge enhancements, with our TRELM model demonstrating superior performance through the integration of knowledge paths and memory.

\noindent\textbf{Relation Extraction}: The TRELM model was evaluated on the TACRED benchmark dataset \citep{tacred}, which includes 42 types of semantic relations. We utilized both micro-averaged and macro-averaged metrics to assess performance. As shown in Table~\ref{tacred_result}, TRELM achieved state-of-the-art performance, confirming the benefits of noise-aware knowledge injection and memory-augmented pre-training for Relation Extraction.

\subsection{Language Understanding Tasks}
TRELM was also tested on the General Language Understanding Evaluation (GLUE) benchmark \cite{wang2018GLUE}. According to the results in Table~\ref{tab:glue}, TRELM slightly outperforms RoBERTa and shows an average improvement of 0.4\% over CoLAKE. Overall, the experiments validate TRELM's marked enhancement in knowledge-aware tasks and its competitive edge in general natural language understanding tasks.

\begin{table}
\centering
\begin{small}
\begin{tabular}{l|ccc}
\toprule
Model & Precision & Recall & F1 \\
\midrule
BERT & 67.23$\pm$0.7 & 64.81$\pm$0.6 & 66.00$\pm$0.6 \\
RoBERTa & 70.80$\pm$0.5& 69.60$\pm$0.6& 70.20$\pm$0.5 \\ 
\midrule
ERNIE & 70.01$\pm$0.8 & 66.14$\pm$0.7& 68.09$\pm$0.7 \\
KnowBERT & 71.62$\pm$0.7& 71.49$\pm$0.6& 71.53$\pm$0.8\\ 
DKPLM & 72.61$\pm$0.5& 73.53$\pm$0.4& 73.07$\pm$0.5\\ 
KP-PLM & 72.60$\pm$0.8& 73.70$\pm$0.7 & 73.15$\pm$0.7\\ 
KALM & 72.52$\pm$0.8& 73.38$\pm$0.9& 72.95$\pm$0.8\\ \midrule
TRELM  & \textbf{72.89}$\pm$0.5 &\textbf{73.84}$\pm$0.4 &\textbf{73.36}$\pm$0.4 \\ \bottomrule
\end{tabular}
\end{small}
\caption{Model performance on TACRED (\%).}
\label{tacred_result}
\end{table}

\subsection{Analysis of Pre-training Efficiency}
\begin{figure}[tbp]
\centering
\includegraphics[width=0.975\columnwidth]{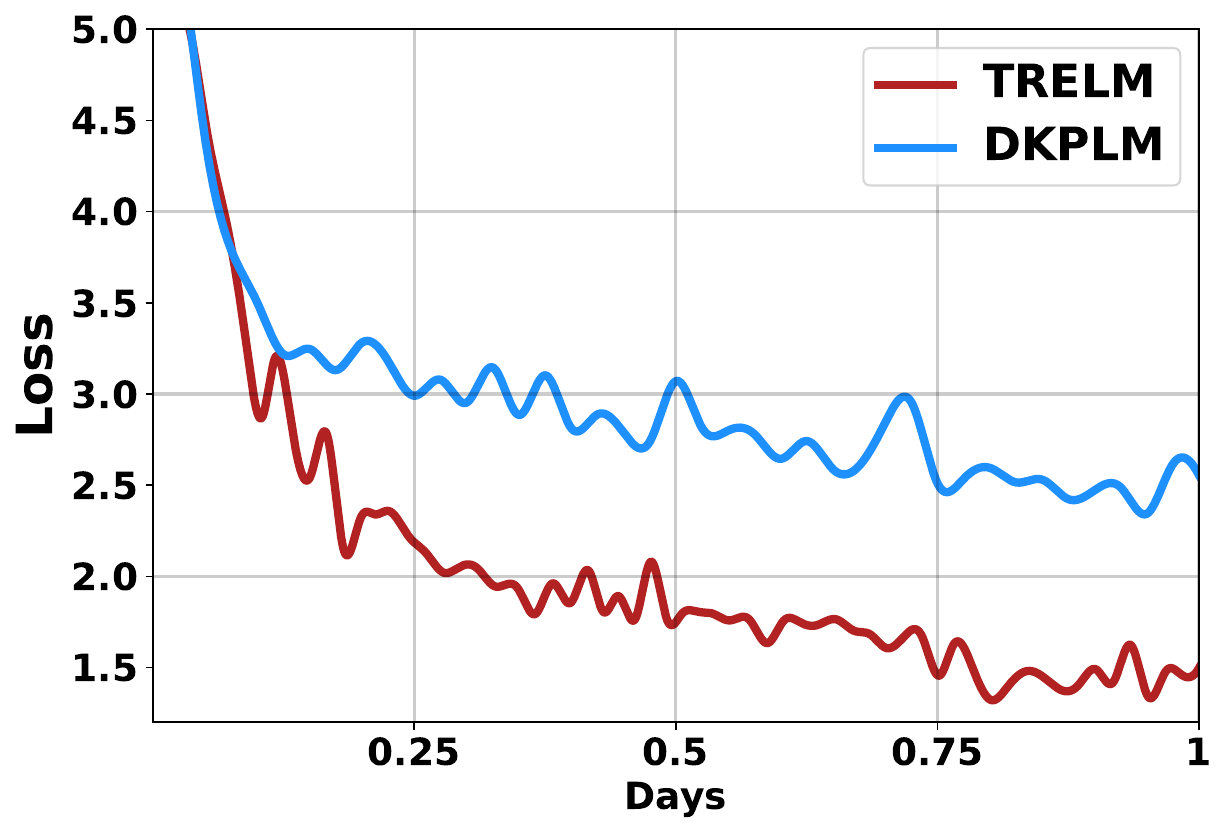}
\caption{The curves of the pre-training loss.}
\label{fig:curve}
\end{figure}

\begin{figure}[tbp]
\centering
\begin{tabular}{ll}
\begin{minipage}[t]{0.47\linewidth}
    \includegraphics[width = 1\linewidth]{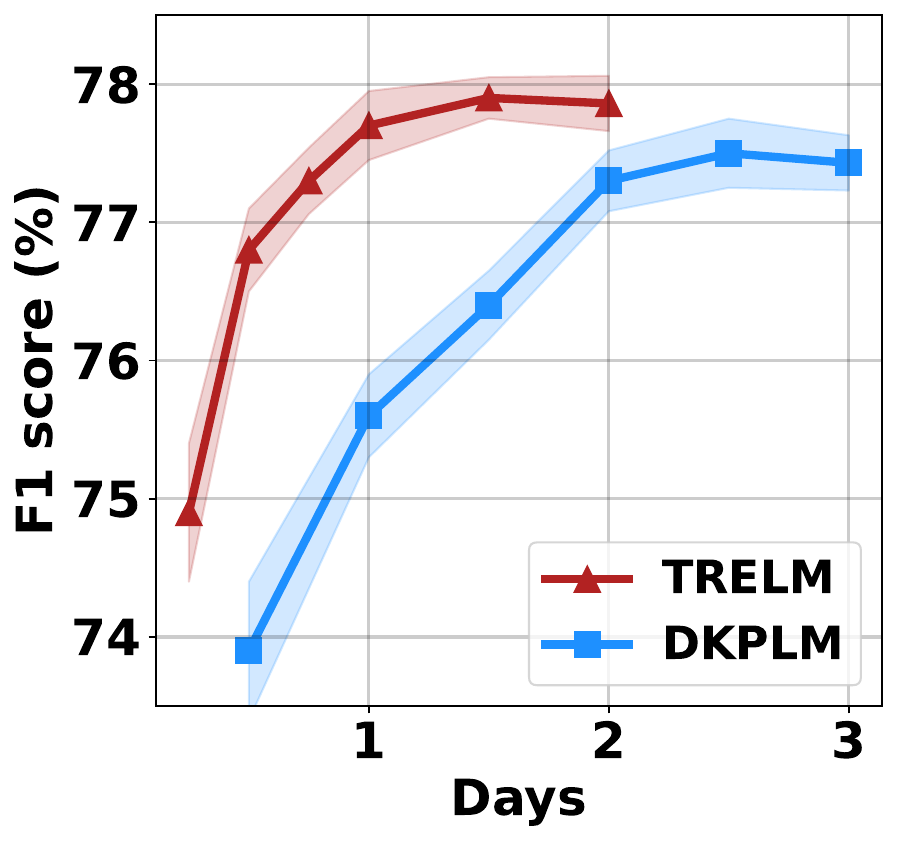}
\end{minipage}
\begin{minipage}[t]{0.48\linewidth}
    \includegraphics[width = 1\linewidth]{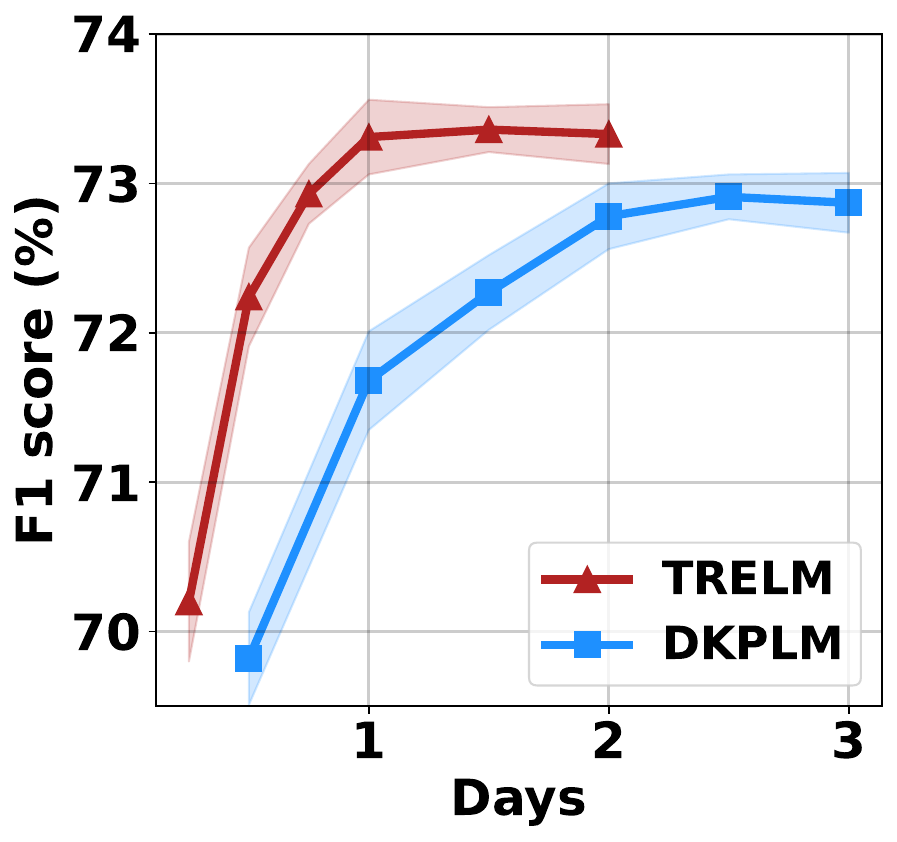}
\end{minipage}
\end{tabular}
\caption{F1 score on Open Entity and TACRED for models trained under the same experiment setting.}
\label{fig:comparison}
\end{figure}

Pre-training was conducted on a server equipped with eight NVIDIA Tesla A100-80G GPUs for both TRELM and DKPLM to ensure a fair comparison. The pre-training loss and F1 scores on Open Entity and TACRED, as illustrated in Figure~\ref{fig:curve} and Figure~\ref{fig:comparison}, demonstrate the efficiency of both models over time.
As depicted in Figure~\ref{fig:curve}, TRELM's loss converges more rapidly than that of DKPLM, suggesting that the incorporation of a memory bank and dynamic knowledge routing contributes to faster model training. The loss curves for TRELM also exhibit greater smoothness, potentially reflecting the evolving quality of memory bank embeddings with continued training.
By evaluating models using checkpoints saved at intervals of $\{0.25, 0.5, 0.75, 1, 1.5, 2\}$ days for TRELM and $\{0.5, 1, 1.5, 2, 2.5, 3\}$ days for DKPLM, we observe from Figure~\ref{fig:comparison} that TRELM consistently outperforms DKPLM in terms of F1 scores. Notably, TRELM's performance within the first 0.75 days is comparable to that of DKPLM after 2 days, indicating that TRELM requires at least $50\%$ less pre-training time to achieve similar results.
In summary, TRELM reaches convergence in approximately one day, whereas DKPLM necessitates around two days, underscoring TRELM's greater pre-training efficiency.

\begin{figure}[tbp]
\centering
\includegraphics[width=0.975\columnwidth]{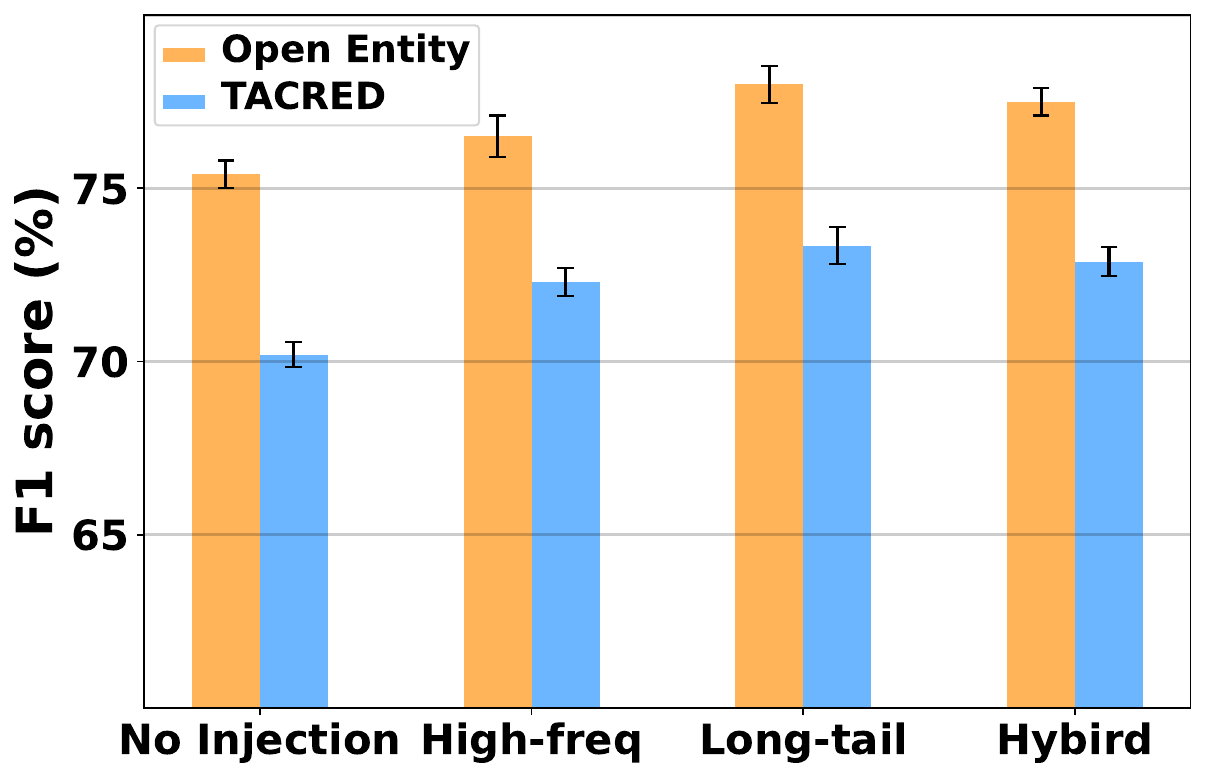}
\caption{Injection method efficiency over Open Entity and TACRED.}
\label{fig:freq}
\end{figure}

\subsection{Influence of Entities with Different Frequencies}
We examined the impact of different knowledge injection strategies on TRELM, focusing on treatments involving only long-tail entities, only high-frequency entities, and a combination of the two. Utilizing the TACRED and Open Entity datasets, we measured the F1 score to assess the effectiveness of our noise-aware knowledge injection method.
Figure~\ref{fig:freq} presents several key insights:
(1) Injecting knowledge into long-tail entities yields better results than limiting it to high-frequency entities, suggesting a greater benefit in enriching representations for entities with sparse occurrences.
(2) Superior performance can be achieved by selectively incorporating knowledge into specific subsets of entities, rather than indiscriminately targeting all available entities.
(3) Our observations are consistent with the findings of \citealp{zhang2021drop} and \citealp{zhang2022dkplm}, which suggest that an overabundance of knowledge injection may detrimentally affect the model's effectiveness.
These findings underscore the advantages of our method and the significance of strategic knowledge selection and injection in enhancing model performance.

\subsection{Ablation Study}
To elucidate the contributions of individual components, we conducted an ablation study and present the findings in Table~\ref{ablation_study}. 
Specifically, the variant labeled ``\textbf{-} Knowledge Injection'' demonstrates a significant decline in the model's ability to comprehend language when noise-aware knowledge injection is removed, underscoring the importance of this feature for enhancing the base PLMs' performance.
Similarly, the ``\textbf{-} Knowledge Routing'' results indicate not only that this component expedites the pre-training process but also that it makes a valuable contribution to the model's overall efficacy. These observations confirm that both knowledge injection and knowledge routing are integral to achieving the superior results.

\subsection{Analysis of Each Module}
\begin{figure}[tbp]
\centering
\begin{tabular}{ll}
\begin{minipage}[t]{0.48\linewidth}
    \includegraphics[width = 1\linewidth]{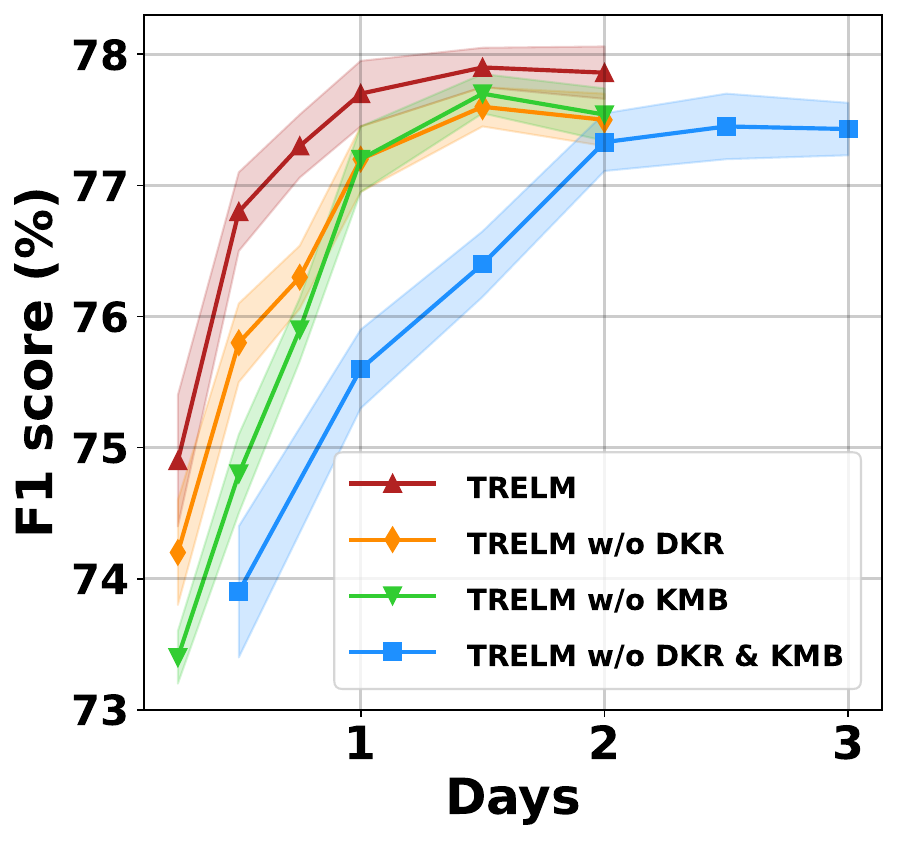}
\end{minipage}
\begin{minipage}[t]{0.48\linewidth}
    \includegraphics[width = 1\linewidth]{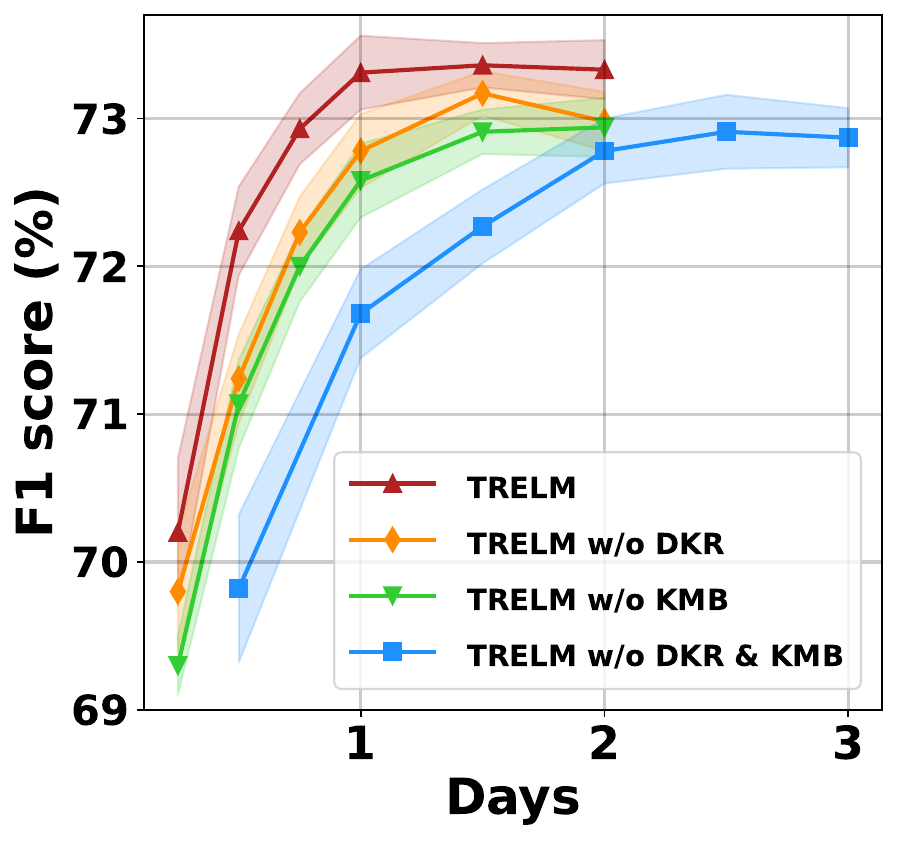}
\end{minipage}
\end{tabular}
\caption{Efficiency of KMB and DKR over Open Entity and TACRED.}
\label{fig:speed}
\end{figure}

In order to understand the individual contributions of each module, we carried out separate experiments to evaluate the specific impact of the Knowledge-augmented Memory Bank (KMB) and the Dynamic Knowledge Routing (DKR) on the pre-training efficiency of TRELM. We focused on quantifying the time saved by each module when used independently. Our analysis of the data illustrated in Figure~\ref{fig:speed} led to the following insights:
(1) Both KMB and DKR enhance the convergence rate of TRELM in the pre-training phase.
(2) KMB exhibits a more pronounced effect on expediting training in the early stages, while DKR's influence becomes increasingly significant over time, ultimately contributing to a greater overall efficiency. This trend may be attributed to an initial period where knowledge pathways are not yet fully established. As the model's capability to accurately assign knowledge improves, DKR's role in pinpointing precise knowledge paths intensifies, thereby boosting its contribution to training efficiency.

\begin{table}[tbp]
\centering
\begin{small}
\begin{tabular}{l cc}
\toprule
Model & TACRED & Open Entity \\ \midrule
TRELM & \bf 73.34\% & \bf 78.0\% \\ 
\quad \textbf{-} Knowledge Injection  & 72.35\%  &  76.8\%\\
\quad \textbf{-} Memory Bank  & 72.91\%  &  77.7\%\\
\quad \textbf{-} Knowledge Routing & 73.17\% &  77.6\%  \\ \bottomrule
\end{tabular}
\end{small}
\caption{Ablation study in terms of F1. }
\label{ablation_study}
\end{table}

\vspace{-1em}

\subsection{Hyper-parameter Analysis}
\label{subsec:hyper-anal}
\begin{figure}[tbp]
\centering
\includegraphics[width=0.975\columnwidth]{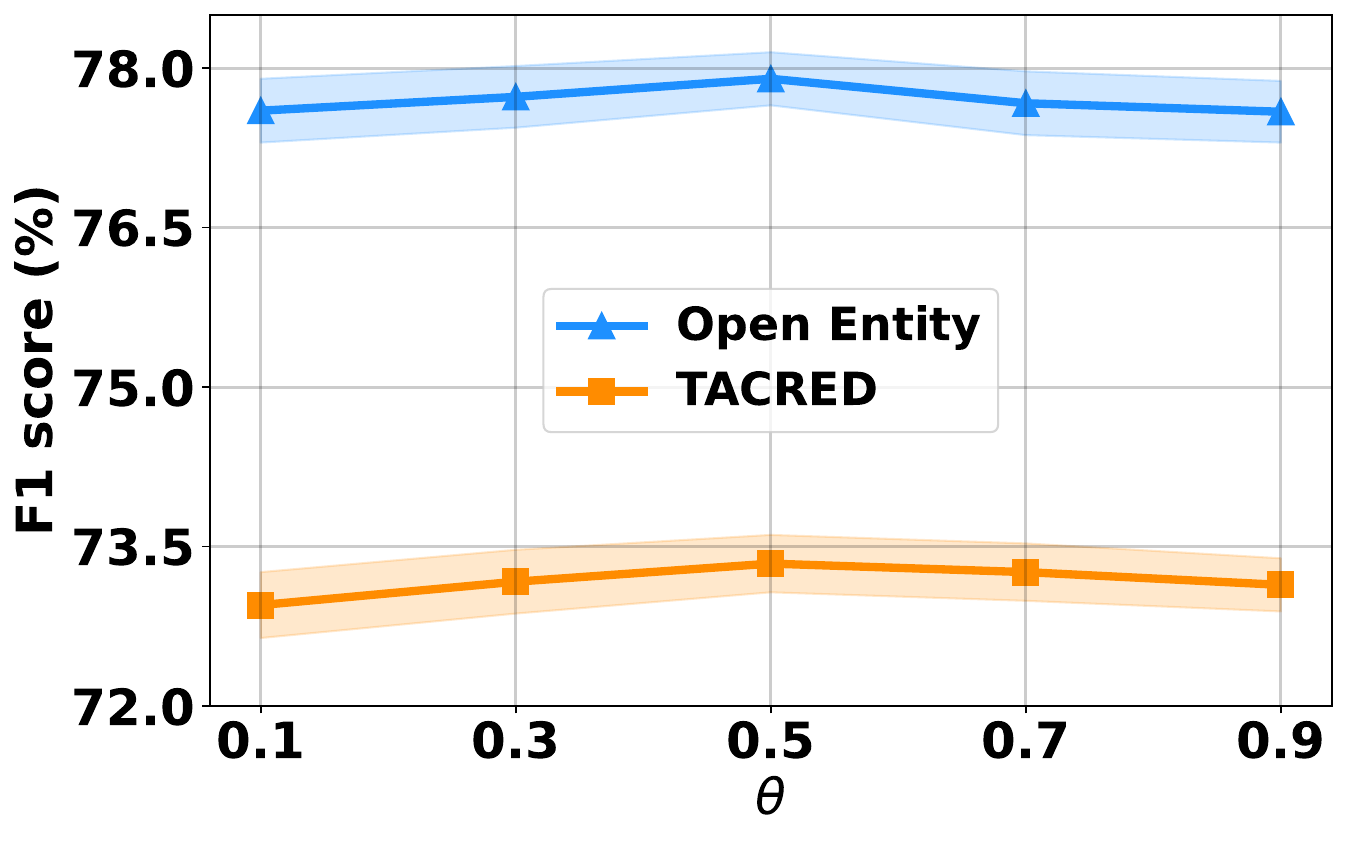}
\caption{Hyper-parameter efficiency of $\theta$ over Open Entity and TACRED.}
\label{fig:hyper_theta}
\end{figure}

\begin{figure}[tbp]
\centering
\begin{tabular}{ll}
\begin{minipage}[t]{0.48\linewidth}
    \includegraphics[width = 1\linewidth]{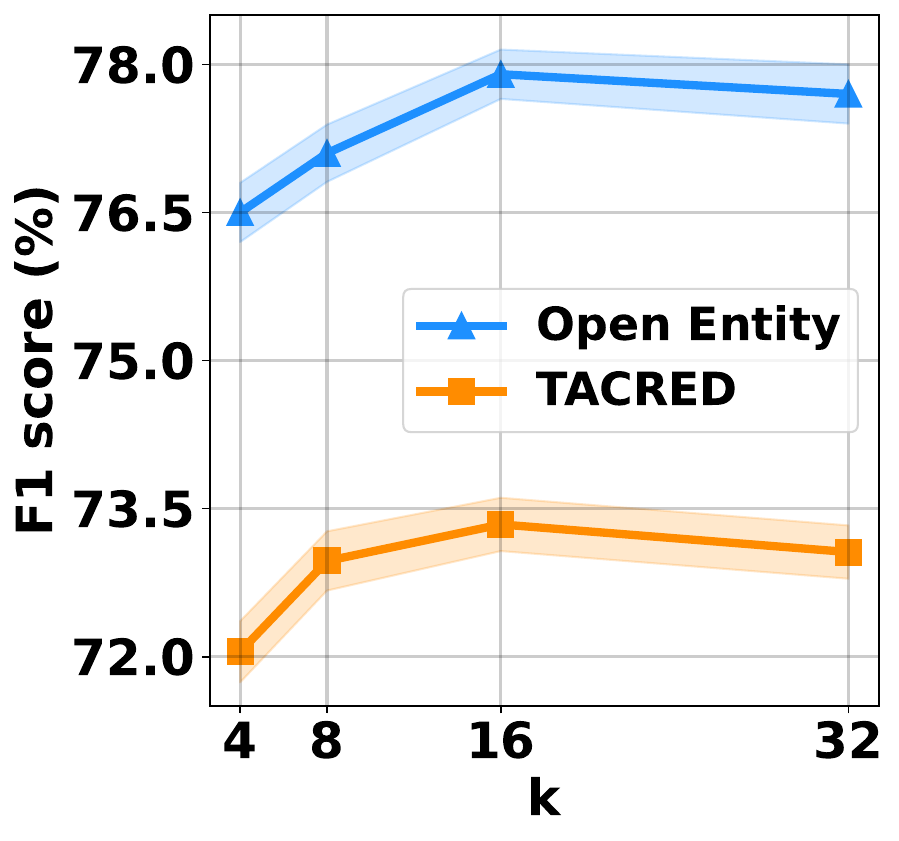}
\end{minipage}
\begin{minipage}[t]{0.48\linewidth}
    \includegraphics[width = 1\linewidth]{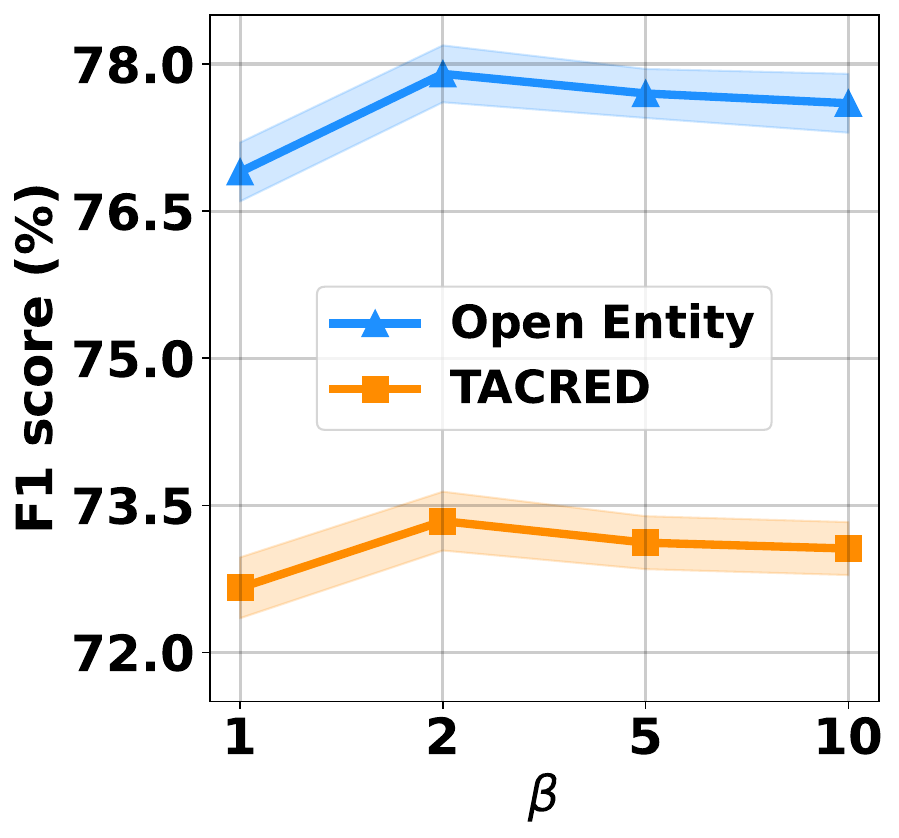}
\end{minipage}
\end{tabular}
\caption{Hyper-parameter efficiency of $k$ and $\beta$  over Open Entity and TACRED.}
\label{fig:hyper_bank}
\end{figure}

We performed a detailed study on the Open Entity and TACRED datasets, focusing on three critical hyper-parameters: the balancing coefficients $\Theta$ for the contrastive knowledge-aware (CKA) task in Eq.~\ref{eq:loss}, the decay rate $\beta$, and the half window size $k$. Each hyper-parameter was varied individually while keeping the others constant.
As shown in Figure~\ref{fig:hyper_theta}, performance initially improves with an increase in $\Theta$, peaks at $\Theta=0.5$, and then diminishes, suggesting an optimal trade-off between the CKA task and other learning objectives at this value.
In Figure~\ref{fig:hyper_bank}, a notable performance boost is observed as the half window size $k$ rises from 4 to 16. However, this upward trend reverses when $k$ is increased to 32, implying that an overly broad context window might introduce irrelevant information that hinders the model's learning.
Referring to the same figure, the lowest performance is seen when $\beta=1$, which corresponds to no decay and a consistent reliance on the memory bank across all pre-training. The model attains its highest performance at $\beta=2$, but further increases in $\beta$ lead to diminishing returns. This decline suggests that a more rapid decrease in the reliance on the memory bank limits the beneficial integration of knowledge, resulting in reduced model efficacy.

\section{Conclusion}
\label{conclusion}
In this paper, we propose TRELM, a robust and efficient training paradigm for pre-training KEPLMs. TRELM introduces two innovative mechanisms designed to streamline the integration of knowledge into PLMs without requiring extra parameters: (1) a knowledge-augmented memory bank that prioritizes knowledge injection for important entities, and (2) a dynamic knowledge routing method that accelerates KEPLMs training and enhances language understanding by updating only the knowledge paths associated with factual knowledge. Our experiments demonstrate that TRELM achieves state-of-the-art performance on knowledge probing tasks and knowledge-aware language understanding tasks. 
For future work, we aim to (1) select more effective knowledge triples from multi-hop neighbor information to inject into KEPLMs, and (2) refine the knowledge paths to further enhance the language understanding capabilities of KEPLMs.

\section*{Acknowledgements}

This work was supported in part by National Natural Science Foundation of China under Grant (No. 62072182) and Alibaba Group through Alibaba Research Intern Program.

 \section*{Limitations}
We have listed some limitations:
1) In the experiments, we only use the WikiData5M as our KG source. We let it as future research to integrate more knowledge sources into PLMs to further improve the performance of downstream tasks.
2) We exclusively trained the model using English data and did not evaluate its performance on low-resource languages, particularly those with limited knowledge graphs. Consequently, the injection of long-tailed entities may not yield significant effectiveness in such scenarios.
3) Our work focuses on the PLMs without Transformer decoders. We think it is possible to extend our method in natural language generation (NLG) tasks.

\section*{Ethical Considerations}
Our contribution in this work is fully methodological, namely a new pre-training paradigm of KEPLMs, achieving the performance improvement of downstream tasks as well as saving the pre-training time. Hence, there is no explicit negative social influences in this work. However, the pre-training data used may have some negative impacts, such as gender and social bias. Our work would unavoidably suffer from these issues. We suggest that users should carefully address potential risks when the models are deployed online.

\section{Bibliographical References}\label{sec:reference}
\bibliographystyle{lrec-coling2024-natbib}
\bibliography{lrec-coling2024-example}

\begin{thebibliography}{0}
\expandafter\ifx\csname natexlab\endcsname\relax\def\natexlab#1{#1}\fi

\end{thebibliography}


\begin{thebibliography}{43}
\expandafter\ifx\csname natexlab\endcsname\relax\def\natexlab#1{#1}\fi

\bibitem[{Broscheit(2019)}]{broscheit2019investigating}
Samuel Broscheit. 2019.
\newblock \href {https://arxiv.org/abs/2003.05473} {Investigating entity knowledge in bert with simple neural end-to-end entity linking}.
\newblock In \emph{Proceedings of the 23rd Conference on Computational Natural Language Learning (CoNLL)}, pages 677--685.

\bibitem[{Cao et~al.(2021)Cao, Lin, Han, Sun, Yan, Liao, Xue, and Xu}]{cao2021kb}
Boxi Cao, Hongyu Lin, Xianpei Han, Le~Sun, Lingyong Yan, Meng Liao, Tong Xue, and Jin Xu. 2021.
\newblock \href {https://doi.org/10.18653/v1/2021.acl-long.146} {Knowledgeable or educated guess? revisiting language models as knowledge bases}.
\newblock In \emph{Proceedings of the 59th Annual Meeting of the Association for Computational Linguistics and the 11th International Joint Conference on Natural Language Processing (Volume 1: Long Papers)}, pages 1860--1874.

\bibitem[{Chang et~al.(2021)Chang, Xu, Xu, and Tu}]{chang2021convolutions}
Tyler Chang, Yifan~Xu Xu, Weijian Xu, and Zhuowen Tu. 2021.
\newblock \href {https://doi.org/10.18653/v1/2021.acl-long.333} {Convolutions and self-attention: Re-interpreting relative positions in pre-trained language models}.
\newblock In \emph{The Joint Conference of the 59th Annual Meeting of the Association for Computational Linguistics and the 11th International Joint Conference on Natural Language Processing}.

\bibitem[{Choi et~al.(2018)Choi, Levy, Choi, and Zettlemoyer}]{openentity}
Eunsol Choi, Omer Levy, Yejin Choi, and Luke Zettlemoyer. 2018.
\newblock \href {https://aclanthology.org/P18-1009/} {Ultra-fine entity typing}.
\newblock In \emph{Proceedings of the 56th Annual Meeting of the Association for Computational Linguistics}, pages 87--96.

\bibitem[{Colon{-}Hernandez et~al.(2021)Colon{-}Hernandez, Havasi, Alonso, Huggins, and Breazeal}]{colon2021combining}
Pedro Colon{-}Hernandez, Catherine Havasi, Jason~B. Alonso, Matthew Huggins, and Cynthia Breazeal. 2021.
\newblock \href {https://arxiv.org/abs/2101.12294} {Combining pre-trained language models and structured knowledge}.
\newblock \emph{CoRR}, abs/2101.12294.

\bibitem[{Cui et~al.(2021)Cui, Cheng, Wu, and Zhang}]{cui2021commonsense}
Leyang Cui, Sijie Cheng, Yu~Wu, and Yue Zhang. 2021.
\newblock \href {https://doi.org/10.18653/v1/2021.findings-acl.61} {On commonsense cues in bert for solving commonsense tasks}.
\newblock In \emph{Findings of the Association for Computational Linguistics: ACL-IJCNLP 2021}, pages 683--693.

\bibitem[{Dai et~al.(2022)Dai, Dong, Hao, Sui, Chang, and Wei}]{dai2022knowledge}
Damai Dai, Li~Dong, Yaru Hao, Zhifang Sui, Baobao Chang, and Furu Wei. 2022.
\newblock \href {https://aclanthology.org/2022.acl-long.581/} {Knowledge neurons in pretrained transformers}.
\newblock In \emph{Proceedings of the 60th Annual Meeting of the Association for Computational Linguistics (Volume 1: Long Papers)}, pages 8493--8502.

\bibitem[{Dong et~al.(2021)Dong, Cordonnier, and Loukas}]{dong2021attention}
Yihe Dong, Jean-Baptiste Cordonnier, and Andreas Loukas. 2021.
\newblock \href {https://doi.org/10.48550/arXiv.2103.03404} {Attention is not all you need: Pure attention loses rank doubly exponentially with depth}.
\newblock In \emph{International Conference on Machine Learning}, pages 2793--2803. PMLR.

\bibitem[{Feng et~al.(2022)Feng, Tan, Zhang, Lei, and Tsvetkov}]{KALM}
Shangbin Feng, Zhaoxuan Tan, Wenqian Zhang, Zhenyu Lei, and Yulia Tsvetkov. 2022.
\newblock \href {https://doi.org/10.48550/arXiv.2210.04105} {{KALM:} knowledge-aware integration of local, document, and global contexts for long document understanding}.
\newblock \emph{CoRR}, abs/2210.04105.

\bibitem[{Ferragina and Scaiella(2010)}]{tagme}
Paolo Ferragina and Ugo Scaiella. 2010.
\newblock \href {https://doi.org/10.1145/1871437.1871689} {Tagme: on-the-fly annotation of short text fragments (by wikipedia entities)}.
\newblock In \emph{Proceedings of the 19th ACM international conference on Information and knowledge management}, pages 1625--1628.

\bibitem[{Han et~al.(2018)Han, Liu, and Sun}]{Neural-knowledge-acquisition}
Xu~Han, Zhiyuan Liu, and Maosong Sun. 2018.
\newblock \href {https://www.aaai.org/ocs/index.php/AAAI/AAAI18/paper/view/16691} {Neural knowledge acquisition via mutual attention between knowledge graph and text}.
\newblock In \emph{Proceedings of the AAAI Conference on Artificial Intelligence}, volume~32.

\bibitem[{Hao et~al.(2021)Hao, Dong, Wei, and Xu}]{attattr}
Yaru Hao, Li~Dong, Furu Wei, and Ke~Xu. 2021.
\newblock \href {https://doi.org/10.1609/aaai.v35i14.17533} {Self-attention attribution: Interpreting information interactions inside transformer}.
\newblock In \emph{Proceedings of the AAAI Conference on Artificial Intelligence}, volume~35, pages 12963--12971.

\bibitem[{He et~al.(2021)He, Zheng, Yang, and Zhang}]{klmo}
Lei He, Suncong Zheng, Tao Yang, and Feng Zhang. 2021.
\newblock \href {https://doi.org/10.18653/v1/2021.findings-emnlp.384} {Klmo: Knowledge graph enhanced pretrained language model with fine-grained relationships}.
\newblock In \emph{Findings of the Association for Computational Linguistics: EMNLP 2021}, pages 4536--4542.

\bibitem[{Kenton and Toutanova(2019)}]{bert}
Jacob Devlin Ming-Wei~Chang Kenton and Lee~Kristina Toutanova. 2019.
\newblock \href {https://doi.org/10.18653/v1/n19-1423} {Bert: Pre-training of deep bidirectional transformers for language understanding}.
\newblock In \emph{Proceedings of NAACL-HLT}, pages 4171--4186.

\bibitem[{Liu et~al.(2020)Liu, Zhou, Zhao, Wang, Ju, Deng, and Wang}]{k-bert}
Weijie Liu, Peng Zhou, Zhe Zhao, Zhiruo Wang, Qi~Ju, Haotang Deng, and Ping Wang. 2020.
\newblock \href {https://ojs.aaai.org/index.php/AAAI/article/view/5681} {K-bert: Enabling language representation with knowledge graph}.
\newblock In \emph{Proceedings of the AAAI Conference on Artificial Intelligence}, volume~34, pages 2901--2908.

\bibitem[{Liu et~al.(2019)Liu, Ott, Goyal, Du, Joshi, Chen, Levy, Lewis, Zettlemoyer, and Stoyanov}]{roberta}
Yinhan Liu, Myle Ott, Naman Goyal, Jingfei Du, Mandar Joshi, Danqi Chen, Omer Levy, Mike Lewis, Luke Zettlemoyer, and Veselin Stoyanov. 2019.
\newblock \href {http://arxiv.org/abs/1907.11692} {Roberta: {A} robustly optimized {BERT} pretraining approach}.
\newblock \emph{CoRR}, abs/1907.11692.

\bibitem[{Peters et~al.(2019)Peters, Neumann, Logan, Schwartz, Joshi, Singh, and Smith}]{knowbert}
Matthew~E Peters, Mark Neumann, Robert Logan, Roy Schwartz, Vidur Joshi, Sameer Singh, and Noah~A Smith. 2019.
\newblock \href {https://doi.org/10.18653/v1/D19-1005} {Knowledge enhanced contextual word representations}.
\newblock In \emph{Conference on Empirical Methods in Natural Language Processing and the 9th International Joint Conference on Natural Language Processing (EMNLP-IJCNLP)}.

\bibitem[{Petroni et~al.(2019)Petroni, Rockt{\"{a}}schel, Riedel, Lewis, Bakhtin, Wu, and Miller}]{lama}
Fabio Petroni, Tim Rockt{\"{a}}schel, Sebastian Riedel, Patrick S.~H. Lewis, Anton Bakhtin, Yuxiang Wu, and Alexander~H. Miller. 2019.
\newblock \href {https://doi.org/10.18653/v1/D19-1250} {Language models as knowledge bases?}
\newblock In \emph{EMNLP}, pages 2463--2473.

\bibitem[{P{\"{o}}rner et~al.(2019)P{\"{o}}rner, Waltinger, and Sch{\"{u}}tze}]{lamauhn}
Nina P{\"{o}}rner, Ulli Waltinger, and Hinrich Sch{\"{u}}tze. 2019.
\newblock \href {http://arxiv.org/abs/1911.03681} {{BERT} is not a knowledge base (yet): Factual knowledge vs. name-based reasoning in unsupervised {QA}}.
\newblock \emph{CoRR}, abs/1911.03681.

\bibitem[{Sun et~al.(2020)Sun, Shao, Qiu, Guo, Hu, Huang, and Zhang}]{colake}
Tianxiang Sun, Yunfan Shao, Xipeng Qiu, Qipeng Guo, Yaru Hu, Xuan-Jing Huang, and Zheng Zhang. 2020.
\newblock \href {https://doi.org/10.18653/v1/2020.coling-main.327} {Colake: Contextualized language and knowledge embedding}.
\newblock In \emph{Proceedings of the 28th International Conference on Computational Linguistics}, pages 3660--3670.

\bibitem[{Sun et~al.(2019)Sun, Wang, Li, Feng, Chen, Zhang, Tian, Zhu, Tian, and Wu}]{baidu-ernie}
Yu~Sun, Shuohuan Wang, Yu{-}Kun Li, Shikun Feng, Xuyi Chen, Han Zhang, Xin Tian, Danxiang Zhu, Hao Tian, and Hua Wu. 2019.
\newblock \href {http://arxiv.org/abs/1904.09223} {{ERNIE:} enhanced representation through knowledge integration}.
\newblock \emph{CoRR}, abs/1904.09223.

\bibitem[{Sundararajan et~al.(2017)Sundararajan, Taly, and Yan}]{aximomatic-attribution}
Mukund Sundararajan, Ankur Taly, and Qiqi Yan. 2017.
\newblock \href {http://proceedings.mlr.press/v70/sundararajan17a.html} {Axiomatic attribution for deep networks}.
\newblock In \emph{International conference on machine learning}, pages 3319--3328. PMLR.

\bibitem[{Wang et~al.(2018)Wang, Singh, Michael, Hill, Levy, and Bowman}]{wang2018GLUE}
Alex Wang, Amanpreet Singh, Julian Michael, Felix Hill, Omer Levy, and Samuel Bowman. 2018.
\newblock \href {https://doi.org/10.18653/v1/w18-5446} {Glue: A multi-task benchmark and analysis platform for natural language understanding}.
\newblock In \emph{Proceedings of the 2018 EMNLP Workshop BlackboxNLP: Analyzing and Interpreting Neural Networks for NLP}, pages 353--355.

\bibitem[{Wang et~al.(2020)Wang, Liu, and Song}]{wang2020language}
Chenguang Wang, Xiao Liu, and Dawn Song. 2020.
\newblock \href {https://arxiv.org/abs/2010.11967} {Language models are open knowledge graphs}.
\newblock \emph{CoRR}, abs/2010.11967.

\bibitem[{Wang et~al.(2022{\natexlab{a}})Wang, Qiu, Zhang, Liu, Li, Wang, Wang, Huang, and Lin}]{DBLP:conf/emnlp/WangQZLLWWHL22}
Chengyu Wang, Minghui Qiu, Taolin Zhang, Tingting Liu, Lei Li, Jianing Wang, Ming Wang, Jun Huang, and Wei Lin. 2022{\natexlab{a}}.
\newblock \href {https://doi.org/10.18653/V1/2022.EMNLP-DEMOS.3} {Easynlp: {A} comprehensive and easy-to-use toolkit for natural language processing}.
\newblock In \emph{Proceedings of the The 2022 Conference on Empirical Methods in Natural Language Processing, {EMNLP} 2022 - System Demonstrations, Abu Dhabi, UAE, December 7-11, 2022}, pages 22--29. Association for Computational Linguistics.

\bibitem[{Wang et~al.(2022{\natexlab{b}})Wang, Huang, Shi, Wang, Qiu, Li, and Gao}]{KP-PLM}
Jianing Wang, Wenkang Huang, Qiuhui Shi, Hongbin Wang, Minghui Qiu, Xiang Li, and Ming Gao. 2022{\natexlab{b}}.
\newblock \href {https://doi.org/10.48550/arXiv.2210.08536} {Knowledge prompting in pre-trained language model for natural language understanding}.
\newblock \emph{arXiv preprint arXiv:2210.08536}.

\bibitem[{Wang et~al.(2021{\natexlab{a}})Wang, Tang, Duan, Wei, Huang, Ji, Cao, Jiang, and Zhou}]{k-adapter}
Ruize Wang, Duyu Tang, Nan Duan, Zhongyu Wei, Xuan-Jing Huang, Jianshu Ji, Guihong Cao, Daxin Jiang, and Ming Zhou. 2021{\natexlab{a}}.
\newblock \href {https://doi.org/10.18653/v1/2021.findings-acl.121} {K-adapter: Infusing knowledge into pre-trained models with adapters}.
\newblock In \emph{Findings of the Association for Computational Linguistics: ACL-IJCNLP 2021}, pages 1405--1418.

\bibitem[{Wang et~al.(2021{\natexlab{b}})Wang, Gao, Zhu, Zhang, Liu, Li, and Tang}]{kepler}
Xiaozhi Wang, Tianyu Gao, Zhaocheng Zhu, Zhengyan Zhang, Zhiyuan Liu, Juanzi Li, and Jian Tang. 2021{\natexlab{b}}.
\newblock \href {https://doi.org/10.1162/tacl\_a\_00360} {{KEPLER:} {A} unified model for knowledge embedding and pre-trained language representation}.
\newblock \emph{Trans. Assoc. Comput. Linguistics}, 9:176--194.

\bibitem[{Wu et~al.(2019)Wu, Fan, Baevski, Dauphin, and Auli}]{wu2019pay}
Felix Wu, Angela Fan, Alexei Baevski, Yann~N. Dauphin, and Michael Auli. 2019.
\newblock \href {https://doi.org/10.48550/arXiv.1901.10430} {Pay less attention with lightweight and dynamic convolutions}.
\newblock In \emph{7th International Conference on Learning Representations, {ICLR} 2019}. OpenReview.net.

\bibitem[{Wu et~al.(2020)Wu, Xing, Li, Ke, He, and Liu}]{TNF}
Qiyu Wu, Chen Xing, Yatao Li, Guolin Ke, Di~He, and Tie{-}Yan Liu. 2020.
\newblock \href {http://arxiv.org/abs/2008.01466} {Taking notes on the fly helps {BERT} pre-training}.
\newblock \emph{CoRR}, abs/2008.01466.

\bibitem[{Xiong et~al.(2020)Xiong, Du, Wang, and Stoyanov}]{xiong2019pretrained}
Wenhan Xiong, Jingfei Du, William~Yang Wang, and Veselin Stoyanov. 2020.
\newblock \href {https://openreview.net/forum?id=BJlzm64tDH} {Pretrained encyclopedia: Weakly supervised knowledge-pretrained language model}.
\newblock In \emph{8th International Conference on Learning Representations, {ICLR} 2020, Addis Ababa, Ethiopia, April 26-30, 2020}. OpenReview.net.

\bibitem[{Xu et~al.(2021)Xu, Guo, Tang, Su, Shou, Gong, Zhong, Quan, Jiang, and Duan}]{xu2021syntax}
Zenan Xu, Daya Guo, Duyu Tang, Qinliang Su, Linjun Shou, Ming Gong, Wanjun Zhong, Xiaojun Quan, Daxin Jiang, and Nan Duan. 2021.
\newblock \href {https://doi.org/10.18653/v1/2021.acl-long.420} {Syntax-enhanced pre-trained model}.
\newblock In \emph{Proceedings of the 59th Annual Meeting of the Association for Computational Linguistics and the 11th International Joint Conference on Natural Language Processing (Volume 1: Long Papers)}, pages 5412--5422.

\bibitem[{Yang and Mitchell(2017)}]{knowledge-bases-lstms}
Bishan Yang and Tom Mitchell. 2017.
\newblock \href {http://arxiv.org/abs/1902.09091} {Leveraging knowledge bases in lstms for improving machine reading}.
\newblock In \emph{Proceedings of the 55th Annual Meeting of the Association for Computational Linguistics (Volume 1: Long Papers)}, pages 1436--1446.

\bibitem[{Ye et~al.(2022)Ye, Zhang, Deng, Chen, Chen, Xiong, Chen, and Chen}]{ye2022ontology}
Hongbin Ye, Ningyu Zhang, Shumin Deng, Xiang Chen, Hui Chen, Feiyu Xiong, Xi~Chen, and Huajun Chen. 2022.
\newblock \href {https://doi.org/10.1145/3485447.3511921} {Ontology-enhanced prompt-tuning for few-shot learning}.
\newblock In \emph{Proceedings of the ACM Web Conference 2022}, pages 778--787.

\bibitem[{Yu et~al.(2022)Yu, Zhu, Yang, and Zeng}]{yu2022jaket}
Donghan Yu, Chenguang Zhu, Yiming Yang, and Michael Zeng. 2022.
\newblock \href {https://ojs.aaai.org/index.php/AAAI/article/view/21417} {Jaket: Joint pre-training of knowledge graph and language understanding}.
\newblock In \emph{Proceedings of the AAAI Conference on Artificial Intelligence}, volume~36, pages 11630--11638.

\bibitem[{Zaremoodi et~al.(2018)Zaremoodi, Buntine, and Haffari}]{Adaptive-knowledge-sharing}
Poorya Zaremoodi, Wray Buntine, and Gholamreza Haffari. 2018.
\newblock \href {https://aclanthology.org/P18-2104/} {Adaptive knowledge sharing in multi-task learning: Improving low-resource neural machine translation}.
\newblock In \emph{Proceedings of the 56th Annual Meeting of the Association for Computational Linguistics (Volume 2: Short Papers)}, pages 656--661.

\bibitem[{Zhang et~al.(2021{\natexlab{a}})Zhang, Deng, Cheng, Chen, Zhang, Zhang, and Chen}]{zhang2021drop}
Ningyu Zhang, Shumin Deng, Xu~Cheng, Xi~Chen, Yichi Zhang, Wei Zhang, and Huajun Chen. 2021{\natexlab{a}}.
\newblock \href {https://doi.org/10.24963/ijcai.2021/552} {Drop redundant, shrink irrelevant: Selective knowledge injection for language pretraining}.
\newblock In \emph{{IJCAI} 2021, Virtual Event / Montreal, Canada, 19-27 August 2021}, pages 4007--4014. ijcai.org.

\bibitem[{Zhang et~al.(2021{\natexlab{b}})Zhang, Cai, Wang, Li, Li, Qiu, Tang, He, and Huang}]{DBLP:conf/cikm/ZhangC0LLQTHH21}
Taolin Zhang, Zerui Cai, Chengyu Wang, Peng Li, Yang Li, Minghui Qiu, Chengguang Tang, Xiaofeng He, and Jun Huang. 2021{\natexlab{b}}.
\newblock \href {https://doi.org/10.1145/3459637.3482436} {{HORNET:} enriching pre-trained language representations with heterogeneous knowledge sources}.
\newblock In \emph{{CIKM} '21: The 30th {ACM} International Conference on Information and Knowledge Management, Virtual Event, Queensland, Australia, November 1 - 5, 2021}, pages 2608--2617. {ACM}.

\bibitem[{Zhang et~al.(2022{\natexlab{a}})Zhang, Dong, Wang, Wang, Wang, Liu, Huang, Li, and He}]{DBLP:conf/emnlp/ZhangDWWWLHLH22}
Taolin Zhang, Junwei Dong, Jianing Wang, Chengyu Wang, Ang Wang, Yinghui Liu, Jun Huang, Yong Li, and Xiaofeng He. 2022{\natexlab{a}}.
\newblock \href {https://doi.org/10.18653/V1/2022.EMNLP-INDUSTRY.57} {Revisiting and advancing chinese natural language understanding with accelerated heterogeneous knowledge pre-training}.
\newblock In \emph{Proceedings of the 2022 Conference on Empirical Methods in Natural Language Processing: {EMNLP} 2022 - Industry Track, Abu Dhabi, UAE, December 7 - 11, 2022}, pages 560--570. Association for Computational Linguistics.

\bibitem[{Zhang et~al.(2022{\natexlab{b}})Zhang, Wang, Hu, Qiu, Tang, He, and Huang}]{zhang2022dkplm}
Taolin Zhang, Chengyu Wang, Nan Hu, Minghui Qiu, Chengguang Tang, Xiaofeng He, and Jun Huang. 2022{\natexlab{b}}.
\newblock \href {https://doi.org/10.1609/aaai.v36i10.21425} {Dkplm: Decomposable knowledge-enhanced pre-trained language model for natural language understanding}.
\newblock In \emph{Proceedings of the AAAI Conference on Artificial Intelligence}, volume~36, pages 11703--11711.

\bibitem[{Zhang et~al.(2023)Zhang, Xu, Wang, Duan, Chen, Qiu, Cheng, He, and Qian}]{DBLP:conf/emnlp/ZhangX0DCQCHQ23}
Taolin Zhang, Ruyao Xu, Chengyu Wang, Zhongjie Duan, Cen Chen, Minghui Qiu, Dawei Cheng, Xiaofeng He, and Weining Qian. 2023.
\newblock \href {https://aclanthology.org/2023.emnlp-main.969} {Learning knowledge-enhanced contextual language representations for domain natural language understanding}.
\newblock In \emph{Proceedings of the 2023 Conference on Empirical Methods in Natural Language Processing, {EMNLP} 2023, Singapore, December 6-10, 2023}, pages 15663--15676. Association for Computational Linguistics.

\bibitem[{Zhang et~al.(2017)Zhang, Zhong, Chen, Angeli, and Manning}]{tacred}
Yuhao Zhang, Victor Zhong, Danqi Chen, Gabor Angeli, and Christopher~D Manning. 2017.
\newblock \href {https://doi.org/10.18653/v1/d17-1004} {Position-aware attention and supervised data improve slot filling}.
\newblock In \emph{Conference on Empirical Methods in Natural Language Processing}.

\bibitem[{Zhang et~al.(2019)Zhang, Han, Liu, Jiang, Sun, and Liu}]{zhang2019ernie}
Zhengyan Zhang, Xu~Han, Zhiyuan Liu, Xin Jiang, Maosong Sun, and Qun Liu. 2019.
\newblock \href {https://aclanthology.org/P19-1139/} {Ernie: Enhanced language representation with informative entities}.
\newblock In \emph{Proceedings of the 57th Annual Meeting of the Association for Computational Linguistics}, pages 1441--1451.

\end{thebibliography}
\bibliographystylelanguageresource{lrec-coling2024-natbib}
\bibliographylanguageresource{languageresource}

\end{document}